# Processes Matter: How ML/GAI Approaches Could Support Open Qualitative Coding of Online Discourse Datasets

John Chen, Alexandros Lotsos, Grace Wang, Lexie Zhao, Bruce Sherin, Uri Wilensky, & Michael Horn
Northwestern University

**Abstract:** Open coding, a key inductive step in qualitative research, discovers and constructs concepts from human datasets. However, capturing extensive and nuanced aspects or "coding moments" can be challenging, especially with large discourse datasets. While some studies explore machine learning (ML)/Generative AI (GAI)'s potential for open coding, few evaluation studies exist. We compare open coding results by five recently published ML/GAI approaches and four human coders, using a dataset of online chat messages around a mobile learning software. Our systematic analysis reveals ML/GAI approaches' strengths and weaknesses, uncovering the complementary potential between humans and AI. Line-by-line AI approaches effectively identify content-based codes, while humans excel in interpreting conversational dynamics. We discussed how embedded analytical processes could shape the results of ML/GAI approaches. Instead of replacing humans in open coding, researchers should integrate AI with and according to their analytical processes, e.g., as parallel co-coders.

## Introduction[1]

Qualitative coding is the process of systematically identifying, generating, and organizing concepts from data. As its first step, open coding is a key inductive approach to discovering and constructing concepts from a dataset (Corbin & Strauss, 2008; Fereday & Muir-Cochrane, 2006). Theorists from both thematic analysis (TA) (Braun & Clarke, 2006) and grounded theory (GT) agree on its goal: to capture as many aspects, patterns, or "coding moments" as possible. However, the criteria of being "open" and "as exhaustive as possible" are often inconsistent in practice (Furniss et al., 2011). Open coding can also be difficult and intensive, particularly for CSCL studies with large discourse datasets.

CSCL researchers have leveraged machine learning (ML) methods for analyzing discourse datasets for decades (Stahl, 2015), particularly for deductive analysis. Yet, open coding is more challenging to support. The number and nature of potential open codes are unknown before the inductive process, making it insufficient to examine with traditional metrics such as inter-rater reliability. While researchers have employed GAI models in open coding (Lopez-Fierro & Nguyen, 2024; Zambrano et al., 2023), without understanding human or machine coders' similarities or differences, it is unclear how AI should be best integrated into open coding processes.

Building on recent studies (Chen, Lotsos, Zhao, Wang, et al., 2024; Lopez-Fierro & Nguyen, 2024; Sinha et al., 2024; Zambrano et al., 2023), each with different ML/GAI approach(es) for open coding, we conducted a systematic study on Physics Lab's online chat channel dataset to evaluate each approach's strengths and weaknesses. The dataset was open-coded by four human coders (three PhD students and one undergraduate intern) and five ML/GAI approaches. After initial analysis, we identified three ML/GAI approaches as more suitable for identifying broader themes (e.g., "community feedback"), while the two item-level approaches produced finer-grained codes closer to the theoretical expectations in open coding (e.g., "invite user feedback"). With open codes from human coders and item-level approaches, we conducted further analysis to understand each's strengths, manifested in the uniquely covered codes that could substantially contribute to the analysis.

With item-level approaches, machine coders impressively identified a vast majority of human codes, capable of interpreting the actions, experiences, or intentions from message contents in online collaboration. Notably, the item-level approach that asked GAI models to use verb phrases identified most human codes while contributing many codes grounded in message contents. Yet, machine coders were less likely to produce codes grounded in conversational dynamics, such as "*change of topic without responding*."

Analytical processes are essential for BOTH human and machine coders to produce high-quality open coding outcomes. We explained much of the machine coders' performances through the analytical processes embedded in their prompts. We suggest 1) researchers should integrate appropriate ML/GAI approaches by matching them with the contextual needs of analytical processes, 2) better ML/GAI approaches for qualitative research may be developed by integrating human coding processes, and 3) instead of replacing humans in the workflow, qualitative researchers should consider using ML/GAI approaches as, and only as, parallel co-coders.

---

[1] This paper was recommended for acceptance as a long paper by CSCL reviewers. However, the conference only allowed us to publish its shorter form. We highly recommend our readers to follow this version for more methodological details.

## Related Literature

<u>Existing ML/GAI Approaches for Open Coding</u>
Qualitative analysis enables in-depth explorations of human experiences by focusing on nuanced interpretations, emotions, and subjective experiences that shape individual and social realities (Rahman, 2016). Thematic Analysis (TA) and Grounded Theory (GT), two widely used qualitative methods, advocate for **open coding** as an essential first step to uncover the underlying reasons and processes that form and transform meanings (Corbin & Strauss, 2008; Rahman, 2016).

ML and GAI techniques have been extensively studied by the CSCL community to support qualitative analysis, particularly for large discourse datasets, yet past studies have mostly focused on deductive analysis. In deductive coding, algorithms aim to mimic humans' results, making way for classification-based methods, such as supervised ML (Zheng et al., 2019) or rule-based systems (Erkens & Janssen, 2008). For open coding, however, the number and nature of underlying codes are unknown before the analysis, necessitating the study of generation approaches. To help with qualitative researchers' open coding processes, recent studies focus on:

1. **Topic modeling**, an unsupervised ML technique, identifies semantically similar word groups (topics) in text-based datasets and has been used for inductive or open coding (Baumer et al., 2020; Cai et al., 2023). The resulting topics help researchers focus on key data patterns. Despite efforts, the difficulty in interpreting and evaluating the results limits its power (Grootendorst, 2022; Sievert & Shirley, 2014).
2. **GAI models** have been more recently adopted for open, inductive coding. By iteratively providing data pieces with relevant instructions (e.g., research questions, coding instructions, desired output format), GAI models can produce codes that humans find more interpretable and useful (De Paoli, 2023; Sinha et al., 2024). However, GAI models can still miss nuance and produce vague themes (De Paoli, 2023; Hamilton et al., 2023) while generating non-grounded results or operating at a coarser level of analysis (Byun et al., 2024; Zambrano et al., 2023). A recent CSCL paper has argued for transparency in researcher-AI collaboration (Lopez-Fierro & Nguyen, 2024). Careful prompt design and prompting strategies are essential (Byun et al., 2024; Sinha et al., 2024), yet few have attempted to compare open coding results from different ML/GAI strategies.

<u>Expectations and Evaluation of Open Coding</u>
While machine deductive coding results have been evaluated by inter-coder reliability (ICR) (de Araujo et al., 2023) or ML metrics such as precision and recall (Rosé et al., 2008), evaluating open codes is more challenging. The challenge stems from the nature of open coding: it should capture as many aspects, patterns, or "codable moments" as possible (Corbin & Strauss, 2008). Moreover, since qualitative researchers have largely adopted non-positivist epistemological stances (Carminati, 2018), there is likely no "ground truth" for reference. While such metrics can still capture consistency between machine and human coders, even complete consistency cannot measure the "openness" or "exhaustiveness" of the codes.

Without a viable alternative, many studies adopt deductive metrics for evaluating machine-generated open codes (Gao et al., 2023; Gebreegziabher et al., 2023; Parfenova, 2024; Rietz & Maedche, 2021). Similarly, many human evaluation studies expect machine coders to match what humans identified (Deiner et al., 2024; Khan et al., 2024). As open codes from machines or humans often use different phrases for the same meaning, a recent paper proposes a "Coverage" metric (Zhao et al., 2024). Measuring semantic similarity, this metric tries to match each generated open code with one from the human results. However, like most other approaches, this metric can only identify cases where machine coders find ideas that human codes identified, which intrinsically underutilizes GAI's potential in identifying novel insights.

Can machine coders identify novel insights to complement human coders? Some exploratory studies seem to agree. A recent CSCL study found that GPT-4 contributed novel subcodes and primed researchers to refine code definitions (Lopez-Fierro & Nguyen, 2024). A study found that ChatGPT-generated codebooks, with human refinement, can achieve higher subjective usability ratings than codebooks by humans alone (Zambrano et al., 2023). Researchers have proposed a computational approach to measure the coverage, density, novelty, and divergence of open codes from multiple coders (Chen, Lotsos, Zhao, Hullman, et al., 2024). Building on team-based coding approaches (Cascio et al., 2019), this approach first merges human and machine coders' open codes into a weighted "reference" list of codes, then calculates the metrics for each coder. The study identifies likely "novel" codes from both machine and human coders, suggesting the potential for ML/GAI approaches to complement human open coding in discourse datasets. Yet, without a systematic evaluation, it is unclear what kind of codes different machine coders could contribute to the analysis.

## Task, Dataset, and Coding Approaches

While researchers often engage in conversations with GAI models to explore their potential for open coding (Lopez-Fierro & Nguyen, 2024; Zambrano et al., 2023), models can react to human directives unexpectedly, making it hard to distinguish GAI and humans' contributions. To understand ML/GAI's capability in identifying emergent insights from online discourses, we build on a recent study, where a text-based dataset was coded by four human coders and five automated coding approaches (Chen, Lotsos, Zhao, Wang, et al., 2024).

The discourse dataset is from the online chat channel of Physics Lab, a mobile learning platform for youths to construct interactive physics simulations and share their projects. Similar online discourse datasets are frequently used in research on CSCL designs. The research question explores "how an online community emerged in Physics Lab." To this end, the dataset contains 127 messages between designers and teachers over the first two months of the community formation. The discourse was split into chunks using signal processing techniques - i.e., finding peaks of timestamp intervals between messages that suggest the boundary between conversations. Each chunk is composed of messages from multiple participants (Table 1).

**Table 1**
***Part of*** *an example chunk of conversation that human and machine coders coded on.*

| User-4235 | Can you also include mechanics experiments? |
|---|---|
| Designer-1 | For example, you can see the corresponding circuit diagram after connecting the physical diagram, or vice versa |
| Designer-1 | Mechanics will have to wait until electromagnetism is figured out; it will take some more time |
| User-4232 | Oh, this is nice |
| Designer-1 | Hope to figure out electromagnetism before the end of the year |
| User-4235 | The 3D effect of your software is very good |
| Designer-1 | With bidirectional conversion, you can directly do problems in the application |
| User-4234 | It's already starting to take shape [Emoji] |
| Designer-1 | For example, see the circuit diagram to connect the physical diagram, or vice versa |
| User-4232 | Don't aim for completeness, it should be categorized and refined one by one |

The dataset was independently open-coded by four human coders. Two have worked with the Physics Lab community for years, bringing in-depth knowledge about the community context. The other two have never worked with it, bringing fewer preconceptions. Three coders are PhD students experienced with open coding, while the fourth is an undergraduate intern who is new to the analysis.

Using the five ML/GAI coding approaches with GPT-4o-0513 (Chen, Lotsos, Zhao, Wang, et al., 2024), Machine coders generated the open codes with assigned roles and tasks, such as "*You are an expert in thematic analysis with grounded theory, working on open coding*." They were also provided with the research question and relevant background information in their prompts. To ensure that machine coding was truly inductive, machine coders received no codebooks or examples beforehand. The entire output can be found in Chen et al. (2024), with 652 machine codes and 340 human codes. We summarize each approach below:

1. **BERTopic + LLM** uses topic modeling (Wallach, 2006), an unsupervised ML technique that identifies groups of semantically similar words (i.e., topics) and explains the topics with LLMs (Grootendorst, 2022). Since BERTopic's instruction was intended for general-purpose labeling, we adapted the LLM prompt to include the same instructions about the research question and context for the study.
2. **Chunk Level** asks LLMs to identify open codes from data chunks (e.g., turns of talk in a conversation, parts of an interview, etc). Variants of this approach have been adopted by recent papers (Lopez-Fierro & Nguyen, 2024; Zambrano et al., 2023).
3. **Chunk Level Structured** is mostly identical to Chunk Level, except that it asks LLMs to generate more than one level of concepts ("categories" or "themes" and "codes" or "subcodes.") Recent papers have started to adopt this approach (Chen, Lotsos, Zhao, Wang, et al., 2024; Sinha et al., 2024)
4. **Item Level** asks LLMs to conduct line-by-line coding as grounded theorists suggest (Gibbs, 2007). Recent work has followed this approach to generate one (Sinha et al., 2024) or multiple codes (Chen, Lotsos, Zhao, Wang, et al., 2024) per line (or per item/message in our case).
5. **Item Level with Verb Phrases** builds on the previous approach but instructs LLMs to use verb phrases for labels explicitly. The design was inspired by methodological literature of qualitative research (Davis et al., 2020; Saldaña, 2014) and reported by recent work (Chen, Lotsos, Zhao, Wang, et al., 2024).

## Empirical Analysis

Building on recent work measuring open codes, we systematically analyzed human and machine open codes in four stages: 1) a first-pass evaluation, which identifies the two item-level approaches as closest to human coders and narrows down the analysis; 2) systematic merging to identify open codes uniquely covered by human or machine coders; 3) identifying each code's potential contribution to further analysis; 4) interpreting each code's grounding of contribution in relation with the raw data. Combined, our mixed methods analysis provides a detailed picture of each ML/GAI approach's strengths and weaknesses in open coding.

Validation: Groundedness & Broadness

Two researchers conducted a first-pass evaluation to validate machine codes. The first reading identified two areas of interest: **Groundedness**, operationalized as either 1) a direct semantic connection between the code and the data, or 2) the existence of a latent relationship between the code and the data (e.g. "curriculum standard" may apply when the designer and users were talking about "symbols used in recent textbooks"); **Broadness**, referring to whether a code is sufficiently specific to contribute meaningfully to the analysis (e.g. "community interaction" is too broad in the study context, since almost every message is its instance). Then, two researchers independently judged both aspects and fully reconciled their findings.

**Table 2**
*An overview of the open codes, human validation results, and algorithm merging results.*

|  | Topic Modeling + LLM | Chunk-Level | Chunk-Level, Structured | Item-Level | Item-Level, Verb Phrases |
|---|---|---|---|---|---|
| # Codes | 23 | 47 | 60 | 240 | 282 |
| # Ungrounded | 3 (13%) | 2 (4.3%) | 5 (8.3%) | 2 (0.8%) | 0 (0%) |
| # Overly Broad | 2 (8.7%) | 3 (6.4%) | 3 (5.0%) | 6 (2.5%) | 7 (2.5%) |

Table 2 summarizes the findings from this stage. Most machine codes are both grounded and provide at least some information for further analysis. **Topic Modeling + LLM** identified slightly more ungrounded codes, often because of oversized topics: for example, neither LLMs nor human researchers could find an appropriate label for a topic of 34 messages. We also identified overly broad codes, such as "user engagement" - a concept that can cover almost every message in the dataset - from multiple machine coders.

**Table 3**
*Codes for "Mechanics will have to wait until electromagnetism is figured out; it will take some more time."*

| Human Coders | future update, managing expectations, explanation of upcoming features, vague on responses to the feature request, preview of update, respond to feature request |
|---|---|
| BERTopic + LLM | feature prioritization |
| Chunk Level | future plans |
| Chunk Level, Structured | *N/A (not identified as part of any code)* |
| Item-Level | development timeline, feature prioritization, subject specific tools, user feedback |
| Item-Level, Verb Phrases | manage user expectations, explain current focus, set timeline for mechanics experiments |

Table 3 provides example codes of the same message from the example conversation in Table 1. Both **Topic Modeling + LLM** and **chunk-level** approaches identified broad themes such as "feature prioritization" or "future plans," roughly matching some human codes. Yet, their codes lack the nuanced aspects of fine-grained human codes, particularly for actions and details. On the other hand, **item-level approaches** are more capable of finding fine-grained open codes. For example, the code "*manage user expectations*" matches the human code "*managing expectations*"; "*explain current focus*" is consistent with "*explanation of upcoming features*" in the conversational context; "*set timeline for mechanics experiments*" both summarize the message's content and contributes to the idea of "timeline," which has the potential to complement human analyzers.

Since only item-level approaches produced fine-grained open codes close to human results, after internal discussions, we decided to focus on them for the remainder of our analysis.

Merging Codes Via Human-AI Collaboration

Machine and human coders often use different labels for the same concept (e.g., "manage user expectations" vs. "managing expectations"). To identify open codes uniquely identified by each coder, we systematically merged open codes from human (340 codes) and machine coders (522 codes) through human-AI collaboration.

Following the process outlined by Chen et al. (2024), we conducted a first-pass, algorithmic merging of the codes. In a nutshell, Chen et al.'s algorithm measures semantic similarity between each code's embedding vectors (using a label + a GAI-generated description) and iteratively merges codes with hierarchical clustering. During the process, ad-hoc labels and definitions were iteratively generated based on the "child" codes' labels, definitions, and examples. The process resulted in 315 merged codes.

**Figure 1**
*A merged code "cater to user needs" from "align with user needs" and "catering to user needs."*

| Label | Item Level Verb Phrase | Item Level | 4 Humans |
|---|---|---|---|
| 2. cater to user needs | Covered ▾ align with user needs | Not covered ▾ user expectation management | Covered ▾ catering to user needs |

"A designer aligns and emphasizes catering to user needs in software design."

To refine the merging of the codes, we designed an interactive interface to support human analysts in finding unique and overlapping codes. For example, the matching algorithm did not find a match for the merged label "cater to user needs" in the item-level approach. When that happens, the interface helps our analysts find possible matches by displaying semantically close item-level codes, e.g., "*user expectation management.*"

Using the interface, two researchers analyzed 81 randomly ordered merged codes to determine which codebook covered the concept as expressed by the merged code. To avoid preconceptions from the algorithm, two researchers independently coded the same codes **without seeing the algorithm's decisions** (i.e., all options displayed as "not covered"). After initial explorations, we settled on the following operational definition: a coding approach is considered to **cover the merged code** only if any of its suggested codes covers 1) exactly the same idea or 2) a narrower scope than the merged code. During the process, both researchers take extensive analytic memos on their rationales and observations.

Two researchers conducted three rounds of analysis. After each round, they calculated the inter-coder reliability and fully reconciled the discrepancies. The Cohen's Kappa was 0.54 (moderate) for codes 1-20; 0.67 (substantial) for codes 21-40; 0.76 (substantial) for codes 41-81; and **0.68 (substantial)** for all 81 codes. The Cohen's Kappa between Researcher 1 and the algorithm is 0.78 (substantial); between Researcher 2 and the algorithm, 0.56 (moderate); and between the consensus and the algorithm, **0.68 (substantial)**. Researcher 1 is familiar with the algorithm, while Researcher 2 is not.

The first row of Table 4 demonstrates our results: out of the 81 human-analyzed codes, 57 codes are uniquely covered by human coders or machine coders under our strict definition. Note that the algorithm does not work under the same definition, but rather a heuristic to "merge codes that could be the same."

Identifying Each Code's Potential for Contribution

With the merged codes, we analyze each code and coder's potential contribution to the analysis. We focused on the following code subsets: codes 1) only identified by humans (17); 2) only identified by one of the machine coders (10 for Item-Level; 20 for Verb Phrase); and 3) only identified by both machine coders (12).

In this stage, the two researchers worked to assess **whether and how each code can conceptually contribute to understanding Physics Lab's community formation**. They wrote analytical memos extensively, attempted to identify countercases, and reconciled their differences. The core idea is the unique contribution of each code: if coders did not identify the code during open coding, were we missing potential insights from the dataset? We categorized each code into **Little**, **Minor**, and **Substantial** Gain.

**Table 4**

*An overview of each coding approach's potential contribution to further analysis.*

|  | Total | 4 Human Coders | Item-Level, Both Approaches | Item-Level | Item-Level, Verb Phrases |
|---|---|---|---|---|---|
| # Uniquely Covered | 57 | 17 | 10 | 10 | 20 |
| - Little Gain | 25 | 7 | 8 | 3 | 7 |
| - Minor Gain | 12 | 2 | 2 | 5 | 3 |
| **- Substantial Gain** | **20** | **8** | **0** | **2** | **10** |

Table 4 summarizes the results of this analysis. A significant portion of uniquely covered codes (25 out of 57, 44%) has **little potential to contribute (Little Gain)**. In most cases (20), this is due to another coder identifying similar or more narrow ideas. For example, both AI coders identified the code "*community context*" from the message "Mainly, the school is building an information-based school." None of the 4 human coders applied an equivalent code. Yet, two coders did each apply a pair of codes, "school needs" and "context," that can capture an equivalent idea in combination. If we remove "*community context*" from the open codes, it would not present an actual loss for the analysis. In other cases, the machine or human code was too vague or confusing to interpret (4), or the machine coder accidentally coded in extra messages humans had ignored (2).

While all other codes have the potential to contribute (**Gain**), we found some codes less likely to form the foundation of a research question or hypothesis. We identified two main scenarios for **Minor Gain** codes: 1) Some codes are too specific for the research question around *community formation*. For example, the Item-Level approach identified the codes of "Augmented Reality" and "Multi-language Support" from a version update note posted by the designer. It is possible that, depending on the research context and data, such codes could usefully contribute to answering a research question. However, given the two researchers' understanding of the research question and the broader context, we judged that this is unlikely. 2) Some codes do not conceptually match other codes in themselves, yet their contributions overlap with existing codes. For example, the item-level verb phrase approach identified "acknowledge provided resources" in a user response: "I saw the group files, thank you." While human coders did not find the exact code, their codes "acknowledgement" or "sending resources" cover a similar conceptual space. Missing the code may lead to potential limited (thus minor) loss.

With this analysis, we found human coders identified 8 unique codes that can contribute substantially to the analysis, while machine coders contributed 12 codes (2 for item-level, 10 for item-level verb phrases).

Understanding the Sources of Substantial Gains

To reveal human and machine coders' relative strengths and weaknesses, our final analysis focuses on the 20 **Substantial Gain** codes identified above. Two researchers first examined each code independently and wrote analytical memos. Then, they coded the codes together. Due to page constraints, we report the most prominent feature of the analysis: the **source**, or grounding, of each coder's **Substantial Gain** codes.

Some codes are grounded in the **content** of the messages it was applied to, where "content" here is narrowly construed. For example, consider this designer message: "Consulting the teachers in the group: which type of intersection is used in the circuit diagrams in the current textbooks?" The item-level verb phrase approach applied a code "*consult on educational standards*." Here, the code is dependent on a narrow focus on the specific subject, "current textbook," as present in the message. Yet, instead of focusing on the subject alone, the machine coder interpreted it as "*educational standards*," a **conversation topic** that can contribute to forming research questions. Perhaps teachers' participation in the chat channel was in part triggered by discussions about educational standards, contributing to community formation. The idea of "*consulting*" is also interesting, yet it is not our focus here: human coders have identified many "consulting" codes already.

Some codes, though grounded in the narrow content of messages, provide further interpretation beyond the immediate content. For example, the item-level verb phrase approach found the code "*align with educational standards*" from the user message "Yes, the common one is still the old style", which answers the designer's question in the last paragraph. The code's core contribution is the higher-level action of "*aligning*." Essentially, the machine coder interprets the message content: by affirming "the old style" to be "the common one," the user was "*aligning*" their response to the designer - and the software's design - with educational standards.

Other codes are grounded in **conversational dynamics**. By the term, we refer to the form and function of the message within an ongoing conversation rather than its content. Take a simple example: As long as a message serves as a response to a previous message, it can be coded as a "response." In our dataset, right before the start of Table 1's conversation, the designer explained the features of circuit diagrams. Then, a user said, "Can you also include mechanics experiments?" A human coder coded this **incident** as "*topic change without response*." The code was inferred from the place and role of the message in the conversation: a user initiated the change of topic without responding to a prior message. If further analysis reveals many similar topic changes, researchers may ask follow-up questions: Is it an indicator where users gained initiative in the conversation? Or is it a disruptive action that would drive other users away?

Codes grounded in conversational dynamics can provide further interpretation as well. For example, after the designer replied, "Mechanics will have to wait until electromagnetism is figured out; it will take some more time," a user wrote: "Don't aim for completeness, it should be categorized and refined one by one." By applying the code "*understanding designer's situation*," a human coder not only identifies the situation - but also points to the idea of "*understanding*" between community members that is worthy of further exploration. How often do we see participants understand each other, and is it related to their experiences in the community? Many research questions could have stemmed from here, beyond the immediate notion of "*designer's situation*."

**Table 5**

*Sources of Substantial Gain codes for each coding approach.*

|  | Total | 4 Human Coders | Item-Level, Both Approaches | Item-Level | Item-Level, Verb Phrases |
|---|---|---|---|---|---|
| # Substantial Gain | 20 | 8 | 0 | 2 | 10 |
| - From Content | 13 | 2 | 0 | 2 | 9 |
| - From Conversational Dynamics | 7 | 6 | 0 | 0 | 1 |

Table 5 presents a summary of this analysis. Machine coders' substantial contributions are concentrated in codes grounded in the narrow **content** of messages (11 out of 13), while human coders' contributions focus more on codes grounded in **conversational dynamics** (6 out of 7).

## Discussions & Conclusions

## Processes Embedded in ML/GAI Coding Approaches

By replicating and comparing recently published ML/GAI coding approaches (Chen, Lotsos, Zhao, Wang, et al., 2024; Grootendorst, 2022; Zambrano et al., 2023), we identified major differences between their outcomes. To maximize ML/GAI's potential for analyzing discourse datasets, we need to understand the analytical processes embedded in those approaches and match them with the contextual needs of human analytical processes.

For each machine coding approach, we sent the same research question (how did Physics Lab's online community emerge), contextual information (the dataset came from the beginning of the software's online chat channel), and instruction of role-playing (you are an expert in thematic analysis with grounded theory, working on open coding). Still, we saw significant differences: 1) numerically, item-level approaches identify far more codes than others, closer to human coders; 2) qualitatively, BERTopic and chunk-level approaches identified conceptually broader labels closer to **themes**. Whereas item-level approaches found **fine-grained codes** (e.g., "express hope for the process"), closer to what grounded theorists expect from the open coding process (Corbin & Strauss, 2008) and matching results of a recent exploratory paper (Sinha et al., 2024).

To properly situate each approach in human analysis processes, we need to understand the mechanical differences leading to the outcome discrepancy: the coding processes embedded in each ML/GAI approach.

While **BERTopic** involves GAI models for interpreting topics, it is fundamentally a clustering method based on items' semantic differences (not all topic modeling methods do that). Semantic similarity empowers this approach and dictates its limitations. BERTopic can identify open codes, as long as the potential examples are semantically close (e.g., when people talk about similar things or use similar terms). It becomes less capable when the examples can be semantically orthogonal (think about humor, sarcasm) or context-dependent (e.g., when people simply said "yeah"). While the second issue may be mitigated by using chunks or "bi-messages" (similar to the idea of a bi-gram) for clustering, the first issue, we believe, will be more difficult to overcome.

Machine coders with **chunk-level approaches** are instructed to identify codes from entire data chunks (e.g., parts of a conversation). Yet, even with access to conversational contexts, they still generate fewer "open codes," and these often look like **themes** more than **codes**. One of the potential issues lies in the fine-tuning processes of conversational GAI models to match "human preference." While it is unclear if a singular human preference may exist, we can all agree that models must stop generation at some point - similar to what we often hope for some humans. Yet, such training can become a blessing or a curse: in our experiment, GPT-4o finds five codes for a five-message conversation, and nine codes for a 42-message conversation. These numbers look reasonable when finding themes, yet when finding open codes, they can be intrinsically insufficient. Since we can never reliably estimate the number of codes beforehand, the issue is challenging to overcome.

Human coders can face the same issue during open coding. When coding a large chunk of data, there may be the temptation to code a small number of higher-level, thematic codes. This is where line-by-line coding can help: it "forces analytic thinking while keeping you close to the data" (Gibbs, 2007). We saw that the same approach seems to help machine coders. In **item-level approaches**, we observed results similar to what Gibbs (2007) suggests: the line-by-line coding process "forces" models to "construct codes that reflect (participants') experience." Moreover, when models are instructed to use verb phrases for labels, they seem to better reflect the actions, experiences, or even intentions of participants, as seen in codes such as "seek confirmation," "express frustration with current limitations," or "align with user needs."

Our finding shows the profound impact of embedded processes on ML/GAI approaches. However, we do not argue that item-level approaches are intrinsically better. Item-level approaches are better for finding open codes that align with grounded theory or thematic analysis's expectations. Chunk-level approaches can provide researchers with a broad overview of the discourse data, where many fine-grained codes become more of a distraction. Topic modeling can do a good job of finding topics, while GAI models can help explain each topic's meaning. Each approach has a place based on its embedded process, and researchers should adopt appropriate ML/GAI approaches **according to their human analytic processes**.

## Towards Human-AI Collaboration in Open Coding Processes

While our findings suggest the potential of item-level approaches in open coding, human and machine coders operate with fundamentally different mechanisms. By systematically examining the potential contributions of human and machine coders, we identified the relative strengths and weaknesses of item-level approaches. For the open coding of discourse datasets, we suggest that researchers should consider using item-level approaches (particularly the verb phrase variant) **as (and only as) a parallel co-coder**.

Before we proceed further, it is essential to acknowledge the study's limitations. We only examined one dataset, one research question, and a randomly sampled subset of human or machine coders' results. There is no guarantee that our findings will automatically generalize to other datasets, research questions, or contexts.

Our central contribution lies within the **process** we developed to **evaluate** machine coders' strengths and weaknesses in open coding in relation to human coders. By cautiously adopting and collaborating with a computational algorithm for code merging, we were able to efficiently and rigorously identify codes covered (and uniquely covered) by each codebook, enabling further analysis of potential contributions. By categorizing each "contributing" code's grounding, we identified the strengths of human and machine coders. By adopting our evaluation process, future researchers may be able to evaluate machine coders' potential on a broader scale, enabling more efficient, reliable, and rigorous human-AI collaboration in open coding.

Working with an online discourse dataset - a type of dataset often analyzed in CSCL work (Friesen, 2009) - machine coders with line-by-line coding processes performed impressively in grounded theory open coding. Of the 81 human-analyzed codes, machine coders found most codes that human coders identified from message contents, plus several more (e.g., "aligning with educational standard") that can inspire human coders and assist in further analysis. However, they are **relatively weaker** when identifying **novel codes** grounded in **conversation dynamics**. Although we did not mention it earlier, machine coders sometimes did identify human codes, such as "*confirm understanding*"; the problem is, they rarely identified codes that humans did not find.

To understand how our results would be impacted by GAI models' intrinsic randomness, we performed additional analyses not mentioned above, which largely replicated our results. Using the same prompts, we reran our analysis with more GAI models (6 in total), temperatures (a parameter that determines the randomness in generation, 5 in total), and repetitions (with GPT-4o, temperature at 0.5, five runs). Across the 15 runs, we recognized two human "gain" codes grounded in message contents. For example, for the human Substantial Gain code "noting potential bugs," llama3-70b found conceptual similar "warning about potential issues." Still, machine coders cannot find a remotely similar idea to "change of topic without responding" or any other human Substantial Gain codes grounded in conversational dynamics. Even when we tried OpenAI's latest o1-preview model or more explicit prompts (e.g., "please look for conversational dynamics") with the same data, no models or prompts produced more desirable results.

Still, GAI models are evolving quickly, and more elaborate methods for open coding are possible. Our study has demonstrated the potential of embedding human coding processes into machine coding approaches. A better approach may benefit from more sophisticated coding processes into layered prompts. For example, after machine coders produced first-pass codes identifying the actions of participants, we could run another prompt with the data, first-pass codes, and a new instruction, asking machine coders to further examine the intentions and consequences of such actions. Designing such approaches requires a deep knowledge of AI's capabilities, and there is no guarantee that stronger models or better prompts could achieve more.

Our study shows the potential of ML/GAI approaches in assisting open coding of discourse datasets, a data type often generated and studied by CSCL environments (Stahl, 2015). For now, we suggest researchers consider **using ML/GAI approaches as (and only as) a parallel co-coder** in open coding processes. As a research process, open coding is more than producing open codes alone. Researchers must become acquainted with the data (Corbin & Strauss, 2008), which likely requires that they engage in some open coding themselves. Moreover, we want to be careful about *when* researchers access the output of machine coders, lest they be overly influenced by machine results. As parallel co-coders, ML/GAI approaches can contribute to a more complete picture of ideas from discourse participants, complementing humans' strengths in higher-level interpretations. In the future, ML/GAI approaches may also assist in training junior qualitative researchers. More critical thoughts and broader evaluations are needed to maximize its potential and minimize the risks of ML/GAI in open coding.

## References

Baumer, E. P. S., Siedel, D., McDonnell, L., Zhong, J., Sittikul, P., & McGee, M. (2020). Topicalizer: Reframing core concepts in machine learning visualization by co-designing for interpretivist scholarship.

Braun, V., & Clarke, V. (2006). Using thematic analysis in psychology. *Qualitative Research in Psychology*.

Byun, C., Vasicek, P., & Seppi, K. (2024, May 8). Chain of Thought Prompting for Large Language Model-driven Qualitative Analysis. *Proceedings of the 2024 CHI Conference on Human Factors in Computing Systems*. LLMs as Research Tools: CHI 2024.

Cai, Z., Eagan, B., Marquart, C., & Shaffer, D. W. (2023). LSTM Neural Network Assisted Regex Development for Qualitative Coding. *Advances in Quantitative Ethnography*.

Carminati, L. (2018). Generalizability in Qualitative Research: A Tale of Two Traditions. *Qualitative Health Research*, *28*(13), 2094–2101

Cascio, M. A., Lee, E., Vaudrin, N., & Freedman, D. A. (2019). A Team-based Approach to Open Coding: Considerations for Creating Intercoder Consensus. *Field Methods*, *31*(2), 116–130.

Chen, J., Lotsos, A., Zhao, L., Hullman, J., Sherin, B., Wilensky, U., & Horn, M. (2024). *A Computational*

*Method for Measuring "Open Codes" in Qualitative Analysis*. arXiv Preprint.

Chen, J., Lotsos, A., Zhao, L., Wang, G., Wilensky, U., Sherin, B., & Horn, M. (2024). *Prompts Matter: Comparing ML/GAI Approaches for Generating Inductive Qualitative Coding Results*. arXiv Preprint.

Corbin, J., & Strauss, A. (2008). Chapter 10 / Analyzing Data for Concepts. In *Basics of Qualitative Research (3rd ed.): Techniques and Procedures for Developing Grounded Theory*. SAGE Publications, Inc.

Davis, N. R., Vossoughi, S., & Smith, J. F. (2020). Learning from below: A micro-ethnographic account of children's self-determination as sociopolitical and intellectual action. *Learning, Culture and Social Interaction*, *24*, 100373.

de Araujo, A., Papadopoulos, P. M., McKenney, S., & de Jong, T. (2023). Automated coding of student chats, a trans-topic and language approach. *Computers and Education: Artificial Intelligence*, *4*, 100123.

De Paoli, S. (2023). Performing an Inductive Thematic Analysis of Semi-Structured Interviews With a Large Language Model: An Exploration and Provocation on the Limits of the Approach. Social Science Computer Review, 0(0), 1–23.

Deiner, M. S., Honcharov, V., Li, J., Mackey, T. K., Porco, T. C., & Sarkar, U. (2024). Large Language Models Can Enable Inductive Thematic Analysis of a Social Media Corpus in a Single Prompt: Human Validation Study. *JMIR Infodemiology*, *4*(1), e59641.

Erkens, G., & Janssen, J. (2008). Automatic coding of dialogue acts in collaboration protocols. *International Journal of Computer-Supported Collaborative Learning*, *3*(4), 447–470.

Fereday, J., & Muir-Cochrane, E. (2006). Demonstrating Rigor Using Thematic Analysis: A Hybrid Approach of Inductive and Deductive Coding and Theme Development. *International Journal of Qualitative Methods*, *5*(1), 80–92.

Friesen, N. (2009). Genre and CSCL: The form and rhetoric of the online posting. *International Journal of Computer-Supported Collaborative Learning*, *4*, 171–185.

Furniss, D., Blandford, A., & Curzon, P. (2011). Confessions from a grounded theory PhD: Experiences and lessons learnt. *Proceedings of the SIGCHI Conference on Human Factors in Computing Systems*.

Gao, J., Choo, K. T. W., Cao, J., Lee, R. K.-W., & Perrault, S. (2023). CoAIcoder: Examining the Effectiveness of AI-assisted Human-to-Human Collaboration in Qualitative Analysis. *ACM Transactions on Computer-Human Interaction*, *31*(1), 6:1-6:38.

Gebreegziabher, S. A., Zhang, Z., Tang, X., Meng, Y., Glassman, E. L., & Li, T. J.-J. (2023). PaTAT: Human-AI Collaborative Qualitative Coding with Explainable Interactive Rule Synthesis. *Proceedings of the 2023 CHI Conference on Human Factors in Computing Systems*, 1–19.

Gibbs, G. R. (2007). Thematic coding and categorizing. *Analyzing Qualitative Data*, *703*(38–56).

Grootendorst, M. (2022). BERTopic: Neural topic modeling with a class-based TF-IDF procedure. *arXiv Preprint arXiv:2203.05794*.

Hamilton, L., Elliott, D., Quick, A., Smith, S., & Choplin, V. (2023). Exploring the Use of AI in Qualitative Analysis: A Comparative Study of Guaranteed Income Data. *International Journal of Qualitative Methods*, *22*, 16094069231201504.

Khan, A. H., Kegalle, H., D'Silva, R., Watt, N., Whelan-Shamy, D., Ghahremanlou, L., & Magee, L. (2024). *Automating Thematic Analysis: How LLMs Analyse Controversial Topics* (arXiv:2405.06919).

Lopez-Fierro, S., & Nguyen, H. (2024). *Making Human-AI Contributions Transparent in Qualitative Coding*. Proceedings of ISLS Annual Meetings (CSCL).

Parfenova, A. (2024). Automating Qualitative Data Analysis with Large Language Models. *Proceedings of the 62nd Annual Meeting of the Association for Computational Linguistics.*

Rahman, M. S. (2016). The Advantages and Disadvantages of Using Qualitative and Quantitative Approaches and Methods in Language "Testing and Assessment" Research: A Literature Review. *Journal of Education and Learning*, *6*(1), 102.

Rietz, T., & Maedche, A. (2021). Cody: An AI-Based System to Semi-Automate Coding for Qualitative Research. *Proceedings of the 2021 CHI Conference on Human Factors in Computing Systems*, 1–14.

Rosé, C., Wang, Y.-C., Cui, Y., Arguello, J., Stegmann, K., Weinberger, A., & Fischer, F. (2008). Analyzing collaborative learning processes automatically: Exploiting the advances of computational linguistics in computer-supported collaborative learning. *International Journal of CSCL*.

Saldaña, J. (2014). *Coding and analysis strategies*.

Sievert, C., & Shirley, K. (2014). LDAvis: A method for visualizing and interpreting topics. *Proceedings of the Workshop on Interactive Language Learning, Visualization, and Interfaces*, 63–70.

Sinha, R., Solola, I., Nguyen, H., Swanson, H., & Lawrence, L. (2024). The Role of Generative AI in Qualitative Research: GPT-4's Contributions to a Grounded Theory Analysis. *Proceedings of the Symposium on Learning, Design and Technology*, 17–25.

Stahl, G. (2015). A decade of CSCL. *International Journal of Computer-Supported Collaborative Learning*.

Wallach, H. M. (2006). Topic modeling: Beyond bag-of-words. *ICML '06*, 977–984.

Zambrano, A. F., Liu, X., Barany, A., Baker, R. S., Kim, J., & Nasiar, N. (2023). From nCoder to ChatGPT: From Automated Coding to Refining Human Coding. *Advances in Quantitative Ethnography*.

Zhao, F., Yu, F., & Shang, Y. (2024). A New Method Supporting Qualitative Data Analysis Through Prompt Generation for Inductive Coding. *2024 IEEE International Conference on Information Reuse and Integration for Data Science (IRI)*, 164–169.

Zheng, J., Xing, W., & Zhu, G. (2019). Examining sequential patterns of self-and socially shared regulation of STEM learning in a CSCL environment. *Computers & Education*, *136*, 34–48.